%% file: main.tex
\documentclass[]{article} 
\usepackage{amsmath}
\usepackage{booktabs}
\usepackage{algorithm}
\usepackage{algpseudocode}
\usepackage{graphicx}
\usepackage{caption}
\usepackage{subcaption}
\usepackage{amssymb}
\usepackage{xcolor} 
\usepackage[normalem]{ulem} 

\input{format}

\title{Freezing of Gait Prediction using Proactive Agent that Learns from Selected Experience and DDQN Algorithm}
\begin{document}
\maketitle
\thispagestyle{fancy}
\centering
\author{Septian Enggar Sukmana\footnote{eng.sept@gmail.com}, Sang Won Bae\footnote{sbae4@stevens.edu}, Tomohiro Shibata*\footnote{*tom@brain.kyutech.ac.jp}}\\
\thanks{$^1$$^,$$^3$Graduate School of Life Science and Systems Engineering, Kyushu Institute of Technology, Kitakyushu, Japan\\
$^2$Charles V. Schaefer, Jr. School of Engineering and Science, Stevens Institute of Technology, New Jersey, United States of America
}

\abstract{
Freezing of Gait (FOG) is a debilitating motor symptom commonly experienced by individuals with Parkinson’s Disease (PD) which often leads to falls and reduced mobility. Timely and accurate prediction of FOG episodes is essential for enabling proactive interventions through assistive technologies. This study presents a reinforcement learning-based framework designed to identify optimal pre-FOG onset points, thereby extending the prediction horizon for anticipatory cueing systems. The model implements a Double Deep Q-Network (DDQN) architecture enhanced with Prioritized Experience Replay (PER) allowing the agent to focus learning on high-impact experiences and refine its policy. Trained over 9000 episodes with a reward shaping strategy that promotes cautious decision-making, the agent demonstrated robust performance in both subject-dependent and subject-independent evaluations.  The model achieved a prediction horizon of up to 8.72 seconds prior to FOG onset in subject-independent scenarios and 7.89 seconds in subject-dependent settings.  These results highlight the model’s potential for integration into wearable assistive devices, offering timely and personalized interventions to mitigate FOG in PD patients.
}

\input{Authors/1_introduction.tex}
\input{Authors/2_related_literature.tex}
\input{Authors/3_dataset.tex}
\input{Authors/4_method.tex}
\input{Authors/5_result.tex}
\input{Authors/6_conclusion.tex}

\bibliographystyle{plain}
\bibliography{Authors/bibtex}
\clearpage
\appendix
\input{Authors/App_A}
\input{Authors/App_B}
\end{document}

%% file: format.tex
\usepackage{amssymb}
\usepackage{amsmath}
\usepackage{graphicx}
\usepackage{makeidx}
\usepackage{multicol}
\usepackage[export]{adjustbox}
\usepackage[justification=centering]{caption}
\usepackage{chngcntr}
\counterwithout{figure}{section}
\usepackage{float}

\usepackage{fancyhdr}
\usepackage{booktabs}
\let\origtitle\title 
\renewcommand{\title}[1]{\lfoot{\textit{#1}}\origtitle{\textbf{#1}}}
\renewcommand{\sectionmark}[1]{\markboth {}{}}

\date{}

\newcounter{lemma}

\newcounter{theorem}

%% file: Authors/1_introduction.tex
\section{Introduction}
\label{section:Introduction}

Freezing of gait (FOG) is a critical symptom in Parkinson Disease (PD) where patients suddenly become unable to walk. \textcolor{black}{Consequently, early detection and prediction are essential for timely intervention \cite{ShaheraHossain2025103}}. Deep learning models such as CNN-LSTM have shown effectiveness because CNNs capture local spatial features from accelerometer signals while LSTMs model long-term temporal dependencies \cite{ManqiZhang2025114}. More advanced architectures like DeepConvLSTM-Attention leverage attention mechanisms to focus on the most relevant data segments which is crucial for detecting subtle pre-FOG patterns \cite{ShangaiLi2025112}.  \textcolor{black}{However, most prediction methods rely on supervised learning \cite{Kleanthous2020, Ghayvat2024, Koltermann2023}. While prediction windows have been extended beyond 6 seconds \cite{Fu2025}, these approaches often lack flexibility and depend on complete datasets \cite{ParakkalUnni2020, Wang2020, Filtjens2021}.} Extending the prediction horizon dynamically remains a key challenge for enabling more effective cueing systems for Parkinson’s patients \cite{Kondo2024, Ghayvat2024}.

\begin{figure}[h!]
    \centering
    \begin{subfigure}[b]{0.25\textwidth} 
        \includegraphics[width=\textwidth]{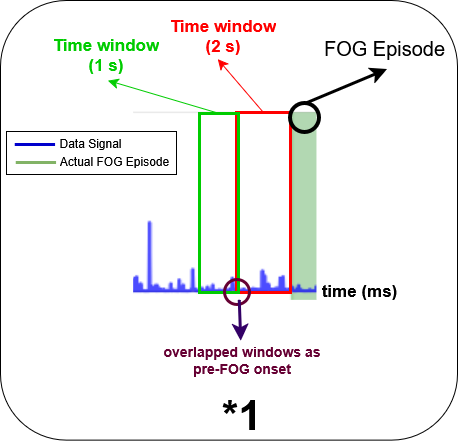} 
        \caption{fixed-window}
        \label{fig:gambar1}
    \end{subfigure}
    \hspace{0.3cm}
    \begin{subfigure}[b]{0.25\textwidth} 
        \includegraphics[width=\textwidth]{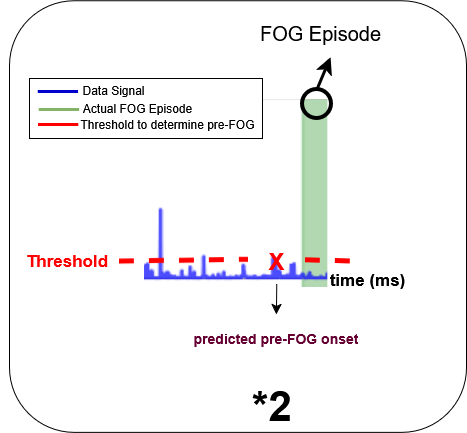} 
        \caption{threshold}
        \label{fig:gambar1}
    \end{subfigure}
    \hspace{0.3cm}
    \begin{subfigure}[b]{0.35\textwidth} 
        \includegraphics[width=\textwidth]{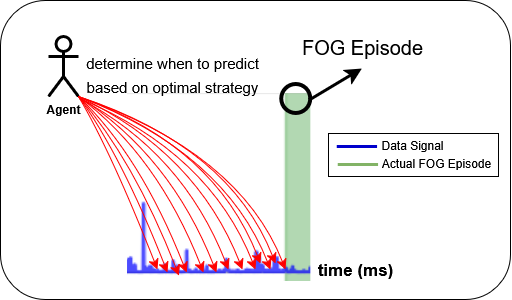} 
        \caption{dynamic-based decision making}
        \label{fig:gambar2}
    \end{subfigure}
    \caption{\textcolor{black}{Illustration of FOG prediction techniques: (a)fixed-window \cite{Naghavi2019,Kleanthous2020,Pardoel2022,Li2023,Pardoel2024}; (b) threshold by machine learning\cite{Fu2025}; (c) proactive using RL agent}}
    \label{fig:reactiveVSproactive}
\end{figure}

To address the limitations of fixed-window supervised learning methods, we propose a proactive agent based on Reinforcement Learning (RL). Unlike prior fixed-window methods, our agent learns when to act to maximize safe cueing time. \textcolor{black}{This approach shifts the paradigm from fixed-window and threshold to dynamic based decision making} (Fig. \ref{fig:reactiveVSproactive}). It allows the agent to decide whether to predict a future FOG event or defer the decision to gather more evidence. Key challenges include high inter-subject variability in pre-FOG patterns and designing a reward function that balances premature predictions (false alarms) against delayed predictions (ineffective alerts). To overcome these issues, we combine Double Deep Q-Networks (DDQN) to mitigate Q-value overestimation with Prioritized Experience Replay (PER). The combination of DDQN and PER is used to emphasize learning from significant experiences while maintaining sample diversity \cite{Seong2024, Fahrmann2022, Ma2021}. The synergy of DDQN and PER has proven effective in complex domains such as financial markets and industrial systems \cite{Ghorrati2024, Parvez2024}, but applying this framework to non-stationary, highly personalized FOG signals introduces unique challenges. These include the ambiguity in defining pre-FOG states and the need for iterative policy refinement based on significant prediction deviations.

By strategically replaying significant experiences while mitigating sampling bias, the agent can recognize a broader spectrum of FOG prediction patterns over extended time horizons and accurately identify optimal intervention points. This paper is organized as follows: Section \ref{section:Introduction} outlines the background and motivation. Section \ref{section: Related literature} reviews related literature. Section \ref{section:Dataset} describes the dataset used in this study. Section \ref{section:Method} presents the methodology, including feature extraction, agent design, and evaluation metrics. Section \ref{section:Result and Discussion} discusses the experimental results and compares them with prior work. Finally, Section \ref{section:Conclusion} summarizes the findings and concludes the paper.

%% file: Authors/2_related_literature.tex
\section{Related literature}
\label{section: Related literature}
Recent advancements in Freezing of Gait (FOG) prediction have implemented several approaches  including statistical analysis of time-series features such as Freeze Index and Sample Entropy \cite{Naghavi2019} as well as the use of fixed-duration pre-FOG transition labels \cite{Kleanthous2020}. These features have served as inputs for conventional machine learning models such SVMs. In parallel, a novel gait-related metrics such as Impaired Gait Features \cite{Zhang2020} and Freeze Index Ratio (FIR) \cite{Li2023} have been proposed to enhance the characterization of gait deterioration preceding FOG episodes. Deep learning techniques particularly CNN is implemented to spectrogram representations of inertial sensor data \cite{El-ziaat2022}. The dynamical systems approaches like Dynamic Mode Decomposition (DMD) \cite{Fu2025} have further expanded the analytical toolkit. However, most of prediction horizons remain constrained to short intervals (typically 1–4 seconds) although there is one who has 6.13 seconds for prediction horizon \cite{Fu2025}. Furthermore, the gap in model efficacy between detection and prediction tasks continues to be observed even when employing multimodal sensor inputs \cite{Pardoel2022}\cite{Pardoel2024}\cite{Shalin2021}.

Although the pre-FOG interval clearly encompasses informative patterns for prediction, current methodological frameworks remain constrained by inherent systemic shortcomings. Key limitations include fixed short-term prediction intervals and dependence on manual feature design \cite{Zhang2020}\cite{Li2023}\cite{El-ziaat2022} which may fail to capture complex temporal dependencies and individual variability. To address these challenges, Reinforcement Learning (RL) has emerged as a promising paradigm. Treating FOG prediction as a decision-making sequence empowers RL agents to refine early warnings and generate longer prediction intervals. This capability is critical for enabling timely activation of cueing interventions, thereby potentially mitigating fall risks and improving quality of life for PD patient.

%% file: Authors/3_dataset.tex
\section{Dataset}
\label{section:Dataset}

This study uses the Daphnet Freezing of Gait Dataset \cite{Daphnet}. It consists of 8 of 10 subjects where FOG is happened when data recording activity is performed. Table \ref{tab:data-subjek} shows the demographic of subjects. Column Age and Disease duration is defined in years, Hoehn-Yahr (H\&Y) is measured in ON condition, and Tested in defines the data recording is conducted at subject ON/OFF condition. Participants performed walking tasks designed to trigger FOG in two sessions: without and with Rhythmic Auditory Stimulation (RAS). Acceleration data were transformed into a Triple Index (TI) representation using Dynamic Mode Decomposition (DMD) \cite{Fu2025}. It decomposes signals into modes $\phi_k$ and amplitudes $\alpha_k$. TI is computed as:
\begin{equation}
a = \max |\alpha_k|,\quad m = \text{mean}_k ||\phi_k||,\quad TI = m \times a
\end{equation}
A decrease in TI indicates a transition from stability to instability preceding a FOG event \cite{Fu2025}. \textcolor{black}{The Triple Index (TI) declines before FoG onset because it captures a reduction in the dominant oscillatory mode amplitudes (a). It reflects a loss of gait coordination and dynamic stability as the patient transitions from normal to pathological movement. This degradation in system dynamics is extracted via DMD which is manifested as a drop in TI values several seconds prior to the clinically observed freeze. }
\begin{table}[h!]
    \centering
    \small
    \caption{Selected Subject on Daphnet in this activity}
    \label{tab:data-subjek}
    \begin{tabular}{c  c  c  p{1.8 cm}  c  c}
        \toprule
        \textbf{ID} & \textbf{Gender} & \textbf{Age} & \textbf{Disease duration} & \textbf{H\&Y} & \textbf{Tested in} \\
        \midrule
        1 & M & 66 & 16 & 3 & OFF \\
        2 & M & 67 & 7 & 2 & ON \\
        3 & M & 59 & 30 & 2.5 & OFF \\
        5 & M & 75 & 6 & 2 & OFF \\
        6 & F & 63 & 22 & 2 & OFF \\
        7 & M & 66 & 2 & 2.5 & OFF \\
        8 & F & 68 & 18 & 4 & ON \\
        9 & M & 73 & 9 & 2 & OFF \\
        \bottomrule
    \end{tabular}
\end{table}

%% file: Authors/4_method.tex
\section{Method}
\label{section:Method}
This section describes a methodology to predict FOG using a Double Deep Q-Network (DDQN) to reduce overestimation bias by decoupling action selection and value evaluation combined with Prioritized Experience Replay (PER) to sample experiences with high TD-error which is corrected by Importance Sampling (IS) weights. Unlike prior work \cite{Fahrmann2022}, our agent is proactive. It is trained to anticipate FOG by maximizing long-term returns. The agent operates within a 15-second pre-FOG window, alternating between waiting and placing a prediction flag. This structured reward system helps the agent learn to make deliberate and accurate predictions over time.

\subsection{Parameter and Setting}
\label{ps}
\textcolor{black}{Six input parameters were extracted from a 10-second window to help agent to understand the complex dynamics of the signal better before it can provide accurate predictions. These parameters are: remaining time $\tau_t$ is used to inform the agent about the timing of predictions, mean $\mu_t$ \cite{Kleanthous2020,Pardoel2022, Naghavi2019, Fu2025, Li2023} used to capture the average value of the signal in the pre-FoG phase, standard deviation $\sigma_t$ \cite{Kleanthous2020, Li2023} can help agents recognize clinical signs that gait is beginning to deteriorate before total freezing occurs, trend $\nabla_t$ (slope) \cite{Zhang2020,Li2023,Fu2025} is used specifically to identify the turning point of gait before freezing occurs, spike score $\psi_t$ helps the agent to detect extreme fluctuations during the pre-FoG phase (\cite{Fu2025}), and Z-score $\zeta_t$ is for faster agent learning convergence. The results of the global ablation implementation without these six parameters are presented in Appendix \ref{section:appendixB}.} 
\begin{equation}
\psi_t = \left| \max(T_{I_{window}}) - \mu_t \right|
\end{equation}

\begin{equation}
\zeta_t = \frac{\max(T_{I_{window}}) - \mu_t}{\sigma_f}
\end{equation}

\begin{figure}[b]
\begin{center}
\includegraphics[width=0.85\textwidth]{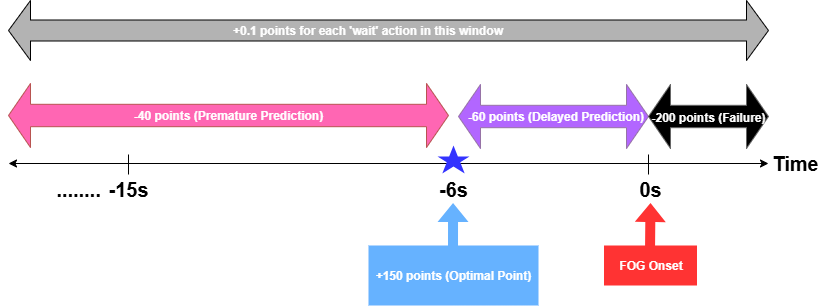}
\caption{\label{RewardDesign} Time-Based Reward Shaping for FOG Prediction}
\end{center}
\end{figure}

\textcolor{black}{Training was conducted over 9,000 episodes with an optimal prediction target of 6 seconds before FOG onset. The reward policy grants +150 for accurate predictions, penalizes early ($-40$ for > 15 s) or late predictions ($-60$ for < 6 s), assigns $-200$ for failure, and provides +0.1 for each `wait' action (Fig \ref{RewardDesign}). This reward scheme represents an improvement over the previous one, as the agent previously faced difficulties in making predictions (Appendix B - Simple Reward Scheme).}

Table \ref{tab:agent_parameters} shows the parameters employed in the agent model. A discount factor ($\gamma$) of 0.99 and a learning rate ($\alpha$) of 0.001 were configured to encourage the agent to prioritize long-term rewards and update its knowledge base in a stable manner. Learning experiences were stored in a replay buffer with a capacity of 50,000 samples. The ReLU (Rectified Linear Unit) activation function was utilized in the hidden layers to process complex data patterns. The evaluation of models performance used the mean squared error (MSE) and its optimization used the Adam optimizer to generate optimal action-value predictions.
\begin{table}[h!]
    \centering
    \small
    \caption{Parameters of the agent model.}
    \begin{tabular}{ll}
        \hline
        \textbf{Parameter} & \textbf{Value} \\
        \hline
        Discount Factor ($\gamma$) & 0.99 \\
        Learning Rate ($\alpha$) & 0.001 \\
        Replay Buffer Size & 50000 \\
        Cost function & Mean Squared Error (MSE) \\
        Optimizer Function & Adam \\
        Activation Function & ReLU (Rectified Linear Unit) \\
        Amount of Hidden Layer & 4 \\
        Amount of Neuron & 256 \\
        \hline
    \end{tabular}
    \label{tab:agent_parameters}
\end{table}

\subsection{Proactive Agent}
\label{pa}
\begin{figure}[b]
\begin{center}
\includegraphics[width=1.0\textwidth]{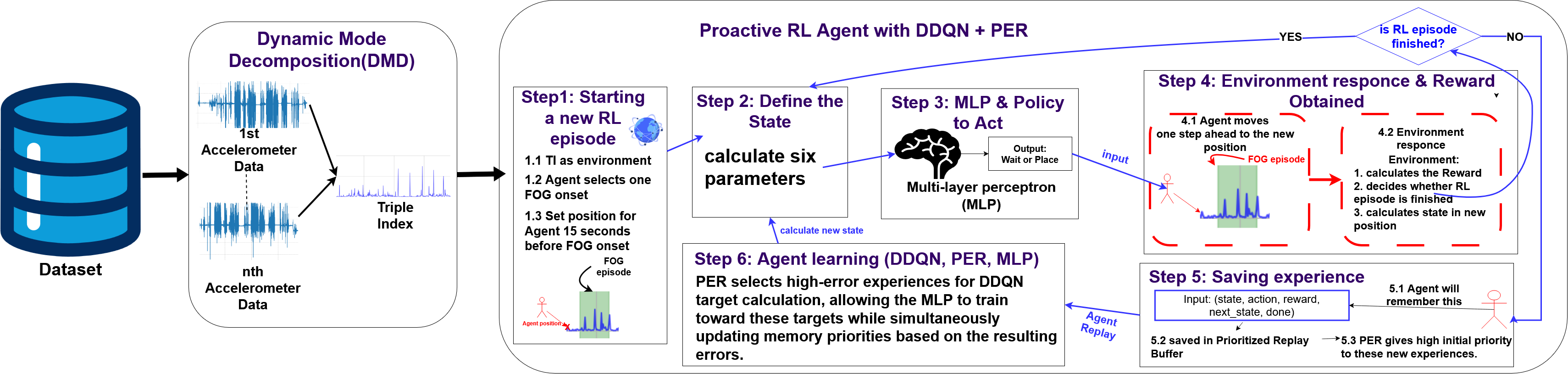}
\caption{\label{fig:flowFull} \textcolor{black}{Proactive agent action to predict FOG}}
\end{center}
\end{figure}

Algorithm \ref{alg:ddqn_per_concise} shows the formulation process. The agent employs a Double Deep Q-Network (DDQN) to approximate the state-action value function $Q(s,a)$ and decide the optimal time to place a flag. At each timestep $t$, the state is represented as a feature vector derived from a recent signal window, including time-to-FOG and five statistical metrics (mean, standard deviation, slope, spike score, z-score) (Fig. \ref{fig:flowFull}). This vector is fed into the Q-network, which selects actions ("wait" or "place flag") using an $\epsilon$-greedy strategy. Policy improvement uses experience replay and DDQN to reduce overestimation bias.
\begin{equation}
a' = \operatorname*{argmax}_{a} Q(s_{t+1}, a), \quad y = r_t + \gamma Q'(s_{t+1}, a')
\end{equation}
To enhance learning efficiency, Prioritized Experience Replay (PER) samples transitions based on TD-error magnitude $|y - Q(s_t, a_t)|$ to emphasize surprising experiences and to correct bias with Importance Sampling weights.

Each training episode focuses on a single pre-FOG group where the environment is initialized by randomly selecting an FOG onsite. Transitions $(s_t, a_t, r_t, s_{t+1}, \text{done})$ are stored in a replay buffer and experience replay is applied by sampling batches of 1024 experiences to train the neural network. As episodes progress, rewards increase and TD-error fluctuations decrease, indicating model convergence through improved best average reward and reduced training loss variation.

\subsection{Internal Baseline (CNN-LSTM)}
\textcolor{black}{The CNN--LSTM is implemented as an internal baseline with two configurations: one using six parameters as utilised by the proposed agent and another without these parameters. The input data are derived from the transformed DaphNet dataset in TI representation with labels binarized as \( y \in \{0,1\} \) where \( y = 1 \) denotes a Pre-FoG window defined as \( T_{\text{pre}} = 6000 \,\text{ms} \) before the onset of motor failure. At each timestep \( t \), a multidimensional feature vector \( x_t \in \mathbb{R}^{6} \) is extracted to create a three-dimensional input tensor \( X \in \mathbb{R}^{N \times L \times F} \) where \( N \) denotes the number of samples, \( L = 50 \) is the sequence length, and \( F = 6 \) is the number of features which is normalized using the standard operator \( z = \frac{x - \bar{x}}{s} \).}

\textcolor{black}{The model architecture consists of spatial-local feature extraction through 1D convolutional layers and max-pooling, followed by temporal dependency modeling using LSTM units to capture contextual transitions toward the Pre-FoG state via gating mechanisms. Model optimization addresses class imbalance by incorporating class weights \( \omega_c \) into the binary cross-entropy loss function.}

\textcolor{black}{The prediction horizon is defined as the interval \( \Delta T = t_{\text{onset}} - t_{\text{alert}} \) where \( t_{\text{alert}} \) corresponds to the first timestep satisfying \( \hat{y}_t > \tau \) within the Pre-FoG window. The threshold \( \tau \) is calibrated to achieve a target False Positive Rate (FPR) of 0.1. The model searches within the interval
\[
T_{\text{window}} = \big[t_{\text{onset}} - 4000 \,\text{ms}, \; t_{\text{onset}}\big]
\]
to determine the earliest activation time \( t_{\text{alert}} = \min \{ t \mid t \in T_{\text{window}} \land A_t = 1 \} \) where
\begin{equation}
A_t =
\begin{cases}
1, & \text{if } \hat{y}_t > \tau, \\
0, & \text{otherwise}.
\end{cases}
\end{equation}
A larger \( \Delta T \) indicates earlier detection of transition patterns toward the Pre-FoG phase, reflecting higher model sensitivity.}

\subsection{Evaluation}
\label{ev}
FOG prediction results are visualized by placing prediction points. The evaluation uses two primary approaches: subject-dependent where training and testing data originate from the same individual (80:20 split) and subject-independent which uses the Leave-One-Subject-Out (LOSO) cross-validation method. Each evaluation reports metrics such as mean reward, mean prediction point, standard deviation, the longest/shortest prediction horizon, the misplaced prediction ratio, and undecided prediction points are incorporated to assess the agent’s accuracy in anticipating FOG onset. \textcolor{black}{The Mean prediction point is calculated as the mean time interval between the agent's 'place flag' action and the actual FOG onset across all successful prediction episodes. Prediction points falling within the FOG episode (misplaced) are excluded from this specific mean to provide a clear measure of the anticipatory window's lead time.} \textcolor{black}{A comparison between the CNN-LSTM model with and without six parameters will also be presented to evaluate the agent's prediction performance.}

\textcolor{black}{The mean prediction horizons of the prior work by Fu et~al. \cite{Fu2025} and the proposed agent are compared to assess FoG prediction performance. Subject-dependent and subject-independent evaluation results are also compared to examine the agent’s generalization and personalization capabilities.} To further assess the agent’s performance in positioning prediction points based on state representations of six parameters, a learning rate curve was employed as a diagnostic tool. This curve aggregates the average reward obtained across reinforcement learning (RL) episodes for all subjects. By analyzing the trajectory of the learning rate curve, the effectiveness of policy updates and the agent’s ability to generalize across varying state conditions can be inferred. 

%% file: Authors/5_result.tex
\section{Result and Discussion}
\label{section:Result and Discussion}

\begin{figure}[h!]
    \begin{center}
    \begin{subfigure}[b]{\linewidth}
    \begin{center}
        \includegraphics[width=0.80\textwidth]{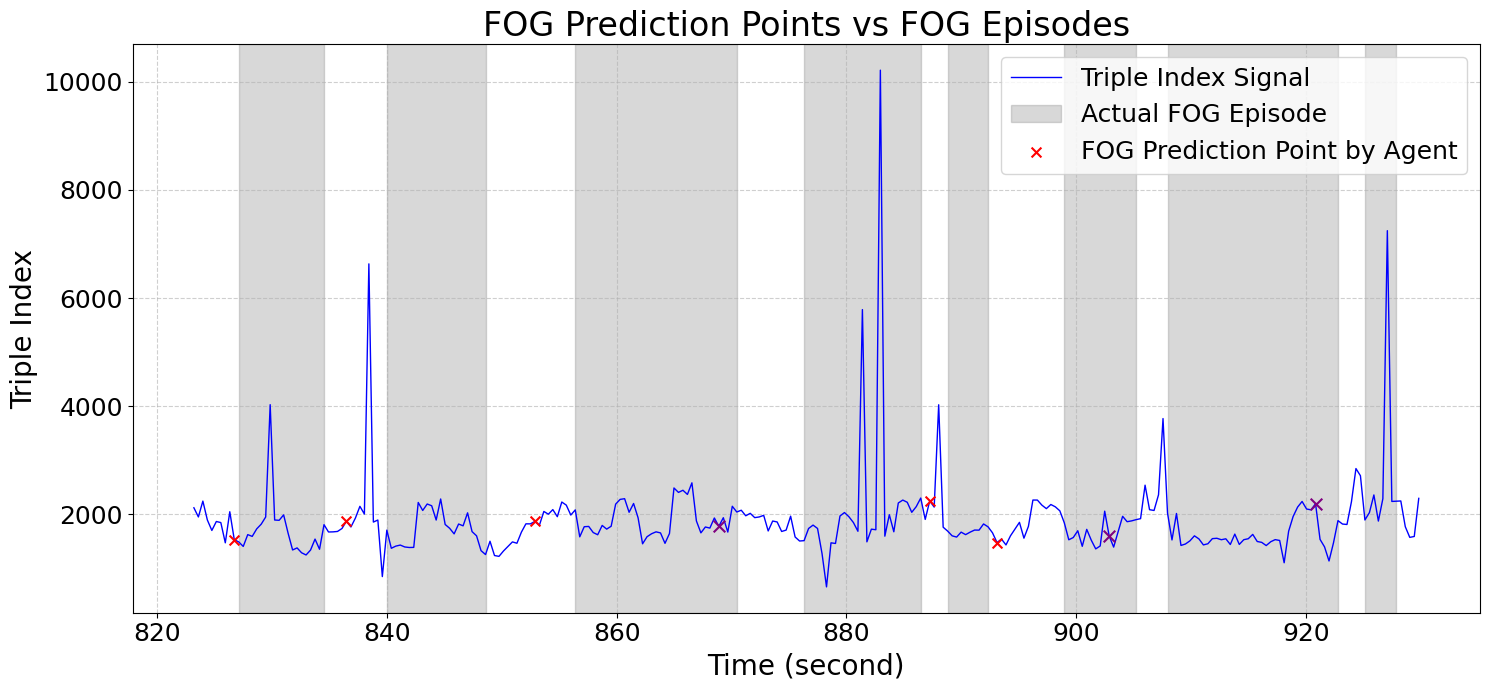}
        \caption{}
        \label{fig:2000eps}
        \end{center}
    \end{subfigure}
    \par\bigskip 
    \begin{subfigure}[b]{\linewidth}
    \begin{center}
        \includegraphics[width=0.80\textwidth]{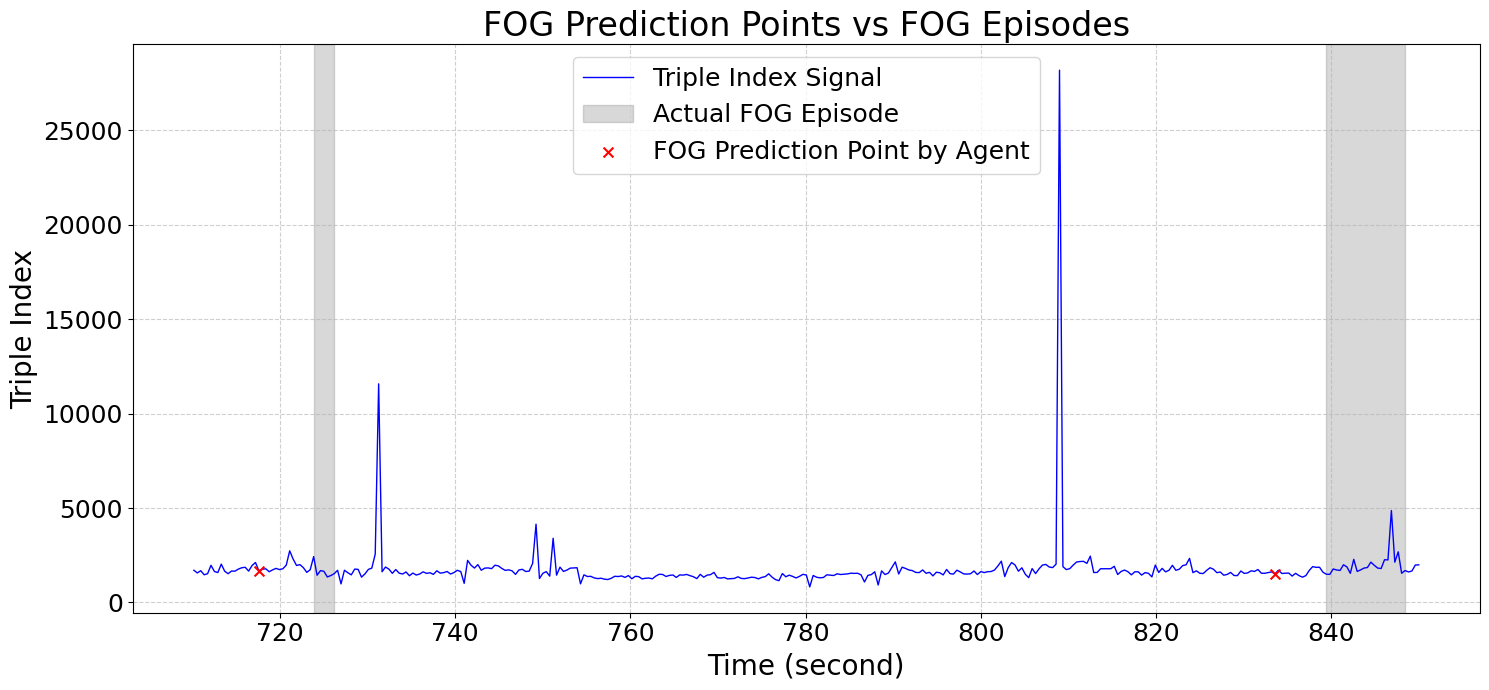}
        \caption{}
        \label{fig:9000eps}
        \end{center}
    \end{subfigure}
    \caption{\textcolor{black}{Example FOG prediction point on: (a) Subject 5; (b) Subject 7}}
    \label{fig:FOG5v7}
    \end{center}
\end{figure}

Figure \ref{fig:FOG5v7} illustrates the agent’s FOG prediction results, showing that most prediction points occurred several seconds before FOG onset. However, some points were misplaced within the FOG episode, likely influenced by the subject’s FOG frequency. \textcolor{black}{Subject 5 who had a higher FOG density exhibited more misplacements than Subject 7, such errors could lead to false alarms in assistive systems.}

To assess these errors, we examined correlations between subject data and misplacement using FOG Density and TI Variance. Spearman analysis revealed a significant positive correlation between FOG Density and Misplaced Ratio in the subject-dependent model ($\rho = 0.76$, \textit{p} = 0.027), while TI Variance showed no effect ($\rho = 0.01$, \textit{p} = 0.97). A similar trend appeared in the subject-independent model ($\rho = 0.63$, \textit{p} = 0.09), though not significant due to small sample size ($N = 8$). These findings suggest FOG Density strongly impacts prediction accuracy, indicating DDQN handles general signal fluctuations well but struggles with closely spaced pre-FOG patterns.

\begin{figure}[b]
\begin{center}
\includegraphics[width=0.55\textwidth]{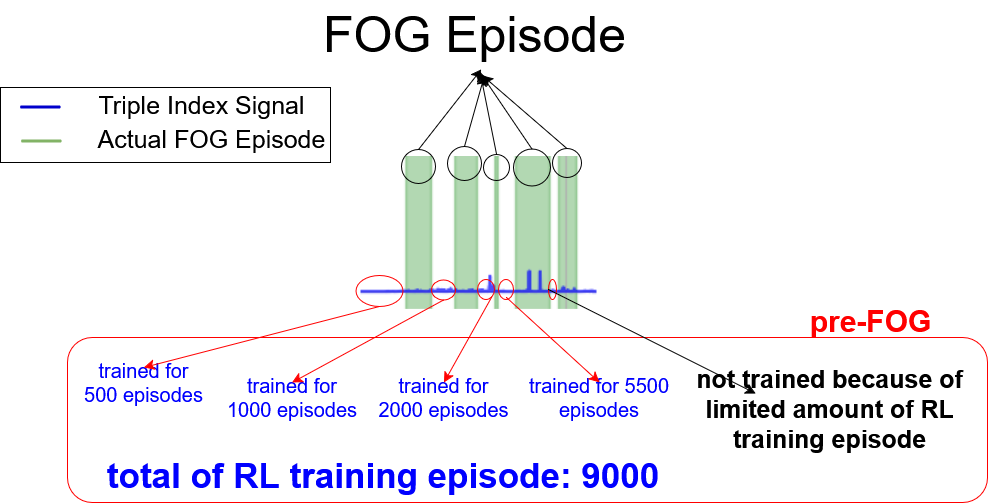}
\caption{\label{fig:FOGDist} \textcolor{black}{Illustration of trained prediction points by replay scheme}}
\end{center}
\end{figure}

Out of the total 307 reported FOG episodes \cite{Daphnet}, 31 episodes were missed by the agent during the subject-independent evaluation and 24 episodes were missed during the subject-dependent evaluation. These missed episodes occurred due to limitations in the RL episodes used during agent training which results in uneven learning distribution (Fig. \ref{fig:FOGDist}). Each pre-FOG segment within a FOG episode exhibits varying patterns that leads to differences in the agent’s intensity of learning for prediction purposes. Consequently, the number of times the agent needs to repeat its learning process for each pre-FOG also varies. \textcolor{black}{Uneven learning distribution in Fig. 5 shows that the limited number of training episodes (9,000) resulted in an imbalanced distribution. Some pre-FoG patterns received intensive training (up to 3,500 episodes) while other segments were not trained at all due to the episode quota constraint.}

\begin{table}[h]
    \centering
    \caption{Experimental Results Based on Subject}
    \label{tab:eksperimen_subjek}
    \resizebox{\textwidth}{!}{%
    \begin{tabular}{l *{8}{c}} 
        \toprule
        \textbf{Variable} & \multicolumn{8}{c}{\textbf{Subject ID}} \\
        \cmidrule(lr){2-9}
        & \textbf{1} & \textbf{2} & \textbf{3} & \textbf{5} & \textbf{6} & \textbf{7} & \textbf{8} & \textbf{9} \\
        \midrule
        \multicolumn{9}{l}{\textbf{Subject Dependent}} \\
        \addlinespace[2pt] 
        Mean - reward & 120.82 & 82.80 & 116.58 & 119.57 & 23.65 & 65.39 & 43.95 & 36.00 \\
        Mean - \textcolor{black}{prediction point (s)} & 5.05 & 3.91 & 6.03 & 5.29 & 2.21 & 3.55 & 3.19 & 2.63 \\
        Std Dev - \textcolor{black}{prediction point (s)} & 1.25 & 1.93 & 1.42 & 1.01 & 1.87 & 1.62 & 0.76 & 1.31 \\
        The longest prediction horizon (s) & 6.87 & 5.08 & 7.89 & 7.29 & 5.74 & 5.47 & 7.26 & 6.70 \\
        The shortest prediction horizon (s) & 4.30 & 1.56 & 4.15 & 4.01 & 3.09 & 1.17 & 5.94 & 3.35 \\
        Misplaced ratio & 0:4 & 0:5 & 1:9 & 4:12 & 0:2 & 0:5 & 0:3 & 0:5 \\
        Undecided prediction point & 0 & 0 & 0 & 0 & 0 & 0 & 0 & 0 \\
        \addlinespace[5pt] 
        \multicolumn{9}{l}{\textbf{Subject Independent}} \\
        \addlinespace[3pt]
        Mean - reward & 116.26 & 91.22 & 81.00 & 88.28 & 112.16 & 100.39 & 16.65 & 94.05 \\
        Mean - \textcolor{black}{prediction point (s)} & 5.64 & 5.63 & 5.38 & 5.15 & 5.04 & 5.50 & 3.49 & 5.44 \\
        Std Dev - \textcolor{black}{prediction point (s)} & 1.39 & 1.31 & 1.96 & 1.67 & 1.56 & 1.63 & 2.59 & 1.12 \\
        The longest prediction horizon (s) & 6.87 & 8.7 & 8.72 & 8.47 & 7.07 & 7.81 & 6.6 & 6.7 \\
        The shortest prediction horizon (s) & 1.62 & 1.95 & 3.32 & 0.45 & 1.33 & 0.78 & 0.66 & 2.39 \\
        Misplaced ratio & 2:18 & 4:20 & 8:39 & 18:51 & 1:10 & 1:21 & 0:13 & 0:24 \\
        Undecided prediction point & 0 & 0 & 1 & 1 & 0 & 0 & 3 & 0 \\
        \bottomrule
    \end{tabular}
    } 
\end{table}

Table \ref{tab:eksperimen_subjek} shows the agent’s performance in placing prediction points under subject-independent and subject-dependent evaluations, where overall higher rewards are observed in the subject-independent setting despite several subjects (3, 5, and 8). However, the reward differences remain relatively small and both evaluations demonstrate stable prediction behavior with mean values consistently exceeding their corresponding standard deviations. Subject 3 achieves the longest prediction horizon for both subject-dependent evaluation (7.89s) and subject-independent evaluation (8.72s). These results exclude the prediction points which are placed within FOG episode (misplaced). Statistical analyses shows that demographic factors do not influence prediction horizon length, as confirmed by the Spearman correlation test for numerical variables (Age, Disease Duration, and H\&Y stage) and the Mann--Whitney U test for categorical variables (Gender and Medication Status) with respect to the longest prediction horizon in both evaluation settings, where no statistically significant correlations or group differences are identified (all $p > 0.25$).

For consistency of pre-FOG patterns across subjects, the patterns appear consistent particularly for Subjects 1, 2, 7, and 9 which show predictions close to the target ($\approx 5.5$--$5.6$~s), low deviation, and minimal error ratios by LOSO evaluation. Subject 9 is ideal (5.44s, Std Dev 1.12, ratio 0:24). In contrast, Subjects 5 and 8 stand out for their inconsistencies where Subject 5 shows an extremely high error ratio (18:51) despite reasonable prediction timing while Subject 8 deviates significantly from the target (3.49s) with the largest deviation (2.59). These findings suggest that while pre-FOG patterns are largely generalizable certain individuals exhibit unique characteristics that hinder cross-subject transferability. \textcolor{black}{We also examined undecided prediction points situation which refers to cases where the agent consistently selects the 'wait' action throughout the entire observation window until reaching the FOG onset without ever executing the 'place' action. Subject-dependent evaluation did not reveal such occurrences. It shows that when the agent is trained specifically on an individual’s data, the agen can reliably recognize gait transition patterns and never misses the opportunity to make a prediction.}

Table \ref{tab:studies} provides a comparative analysis of performance across various studies on FOG prediction tasks. In the evaluation, the agent achieved 5.16s (subject-independent) and 3.98s (subject-dependent) for mean of prediction horizon (SD = 0.77 s; 95\% CI [4.52 \text{ s}, 5.80 \text{ s}], N = 8). \textcolor{black}{In comparison with prior work, Fu et al. \cite{Fu2025} reported a mean prediction horizon of 6.13 ± 0.82 seconds using a DMD-based supervised framework. In the present study, the proposed reinforcement learning agent achieved mean prediction horizons of 5.16 seconds in the subject-independent evaluation and 3.98 seconds in the subject-dependent evaluation. These results indicate that the average prediction horizon of the proposed method does not exceed that reported by Fu et al. \cite{Fu2025} in terms of mean performance.}

\textcolor{black}{However, the key contribution of the proposed approach lies not in improving the global mean prediction horizon but in enabling adaptive and proactive decision-making that yields substantially longer maximum achievable prediction horizons. Specifically, the agent achieved maximum prediction horizons of up to 8.72 seconds in the subject-independent evaluation and 7.89 seconds in the subject-dependent evaluation, exceeding previously reported values. This distinction highlights a fundamental difference between fixed-window supervised approaches and the proposed reinforcement learning framework. While supervised models aim to optimize average performance over predefined windows, the RL agent dynamically determines when to issue a prediction based on the evolving state of gait degradation. As a result, the agent can exploit favorable signal conditions to provide earlier warnings in certain episodes, demonstrating flexibility and personalization that are not captured by mean-based metrics alone. Statistical analyses in this study therefore focus on comparisons between subject-dependent and subject-independent settings, as well as comparisons with the internal CNN–LSTM baseline, rather than direct mean-based superiority over prior studies. These analyses confirm that the proposed agent offers a proactive and adaptive prediction mechanism capable of extending the upper bound of early warning times}. The predictions by the agent also show a longer mean of prediction horizon for both subject-independent and subject-dependent compared to the supervised learning model by implementation of CNN-LSTM as the baseline. While CNN-LSTM is effective at recognizing patterns, it is inherently \textcolor{black}{fixed-window based}. Although this model may be easier to train and interpret \textcolor{black}{also improved after utilizing six parameters (Table \ref{tab:studies})}, it lacks adaptability and flexibility. In contrast, the agent proactively learns the optimal timing for issuing warnings rather than merely matching predefined labels.

\textcolor{black}{To ensure the validity of the parametric analysis, the assumptions of normality and homoscedasticity were tested. The Shapiro-Wilk test confirmed that the prediction horizons for both subject-independent ($p=0.096$) and subject-dependent ($p=0.575$) evaluations followed a normal distribution. Furthermore, Levene’s test was conducted to assess the equality of variances. The results indicated no significant differences in variance between the subject-independent and subject-dependent groups (p=0.944), nor between the agent's performance and the internal baselines (p=1.000). These results statistically justify the application of the t-test for comparing mean prediction horizons.}

\begin{table}[h!]
\centering
\small
\caption{Mean Prediction Horizons from Various Studies}
\label{tab:studies}

    \begin{tabular}{lc} 
        \toprule
        \textbf{\textcolor{black}{Study}} & \textbf{\textcolor{black}{Mean (s)}} \\
        \midrule
        \textcolor{black}{Naghavi and Wade \cite{Naghavi2019}} & \textcolor{black}{2.00} \\
        \textcolor{black}{Zhang et al. \cite{Zhang2020}} & \textcolor{black}{0.93} \\
        \textcolor{black}{Shalin et al. \cite{Shalin2021}} & \textcolor{black}{< 2.00} \\
        \textcolor{black}{El-ziaat et al. \cite{El-ziaat2022}} & \textcolor{black}{1.00} \\
        \textcolor{black}{Pardoel et al. \cite{Pardoel2022}} & \textcolor{black}{0.80} \\
        \textcolor{black}{Pardoel et al. \cite{Pardoel2024}} & \textcolor{black}{0.89} \\
        \textcolor{black}{Fu et al. \cite{Fu2025}} & \textcolor{black}{6.13 $\pm 0.82$} \\
        \textcolor{black}{CNN-LSTM [Baseline without 6 parameters] - Independent} & \textcolor{black}{0.61} \\
        \textcolor{black}{CNN-LSTM [Baseline without 6 parameters] - Dependent} & \textcolor{black}{1.15} \\
        \textcolor{black}{CNN-LSTM [Baseline with 6 parameters] - Independent} & \textcolor{black}{0.97} \\
        \textcolor{black}{CNN-LSTM [Baseline with 6 parameters] - Dependent} & \textcolor{black}{2.77} \\
        \midrule 
        \textcolor{black}{\textbf{This study - Independent}} & \textcolor{black}{\textbf{5.16}} \\
        \textcolor{black}{\textbf{This study - Dependent}} & \textcolor{black}{\textbf{3.98}}  \\
        \bottomrule
    \end{tabular}%
    
\end{table}

\begin{figure}[H]
\begin{center}
\includegraphics[width=0.45\textwidth]{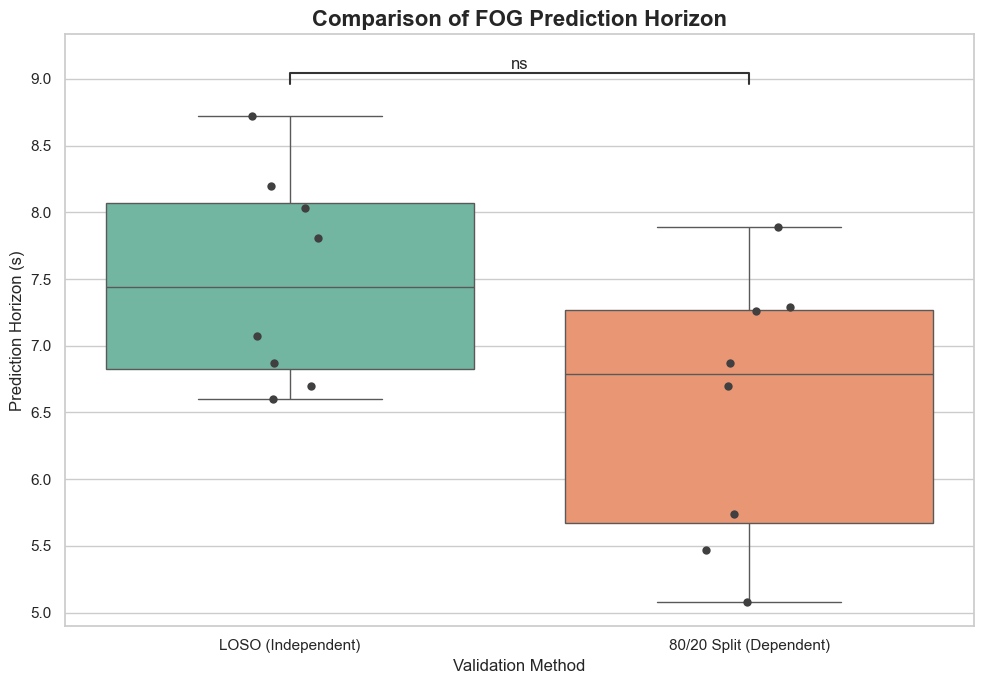}
\caption{\label{fig:losovssplit} Prediction horizon between Subject Independent and Dependent}
\end{center}
\end{figure}

Figure \ref{fig:losovssplit} shows that the LOSO approach tends to perform out and exhibit greater consistency compared to the subject-dependent method (median: 7.4 s vs 6.8 s). However, this difference did not reach statistical significance (p = 0.0506), indicating a strong trend that warrants further validation with a larger sample size.

\begin{figure}[H]
\begin{center}
\includegraphics[width=0.55\textwidth]{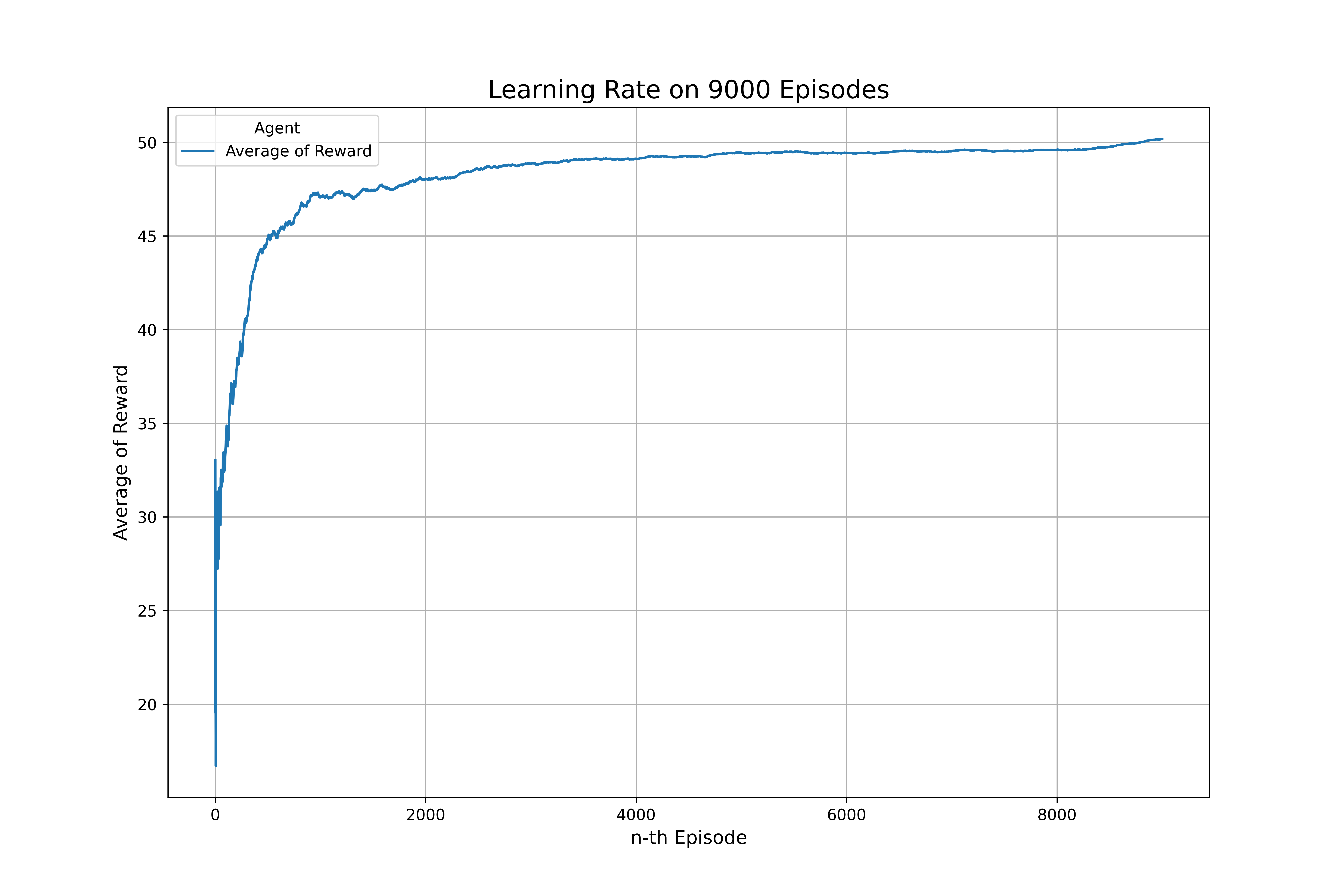}
\caption{\label{fig:learning} Agent learning rate}
\end{center}
\end{figure}

Figure \ref{fig:learning} shows the agent’s learning curve over 0–9000 training episodes. It reflects the agent ability to optimize reward for accurate prediction point placement in FOG episodes. The gradual flattening of the curve illustrates the diminishing returns effect where learning becomes more incremental as the agent transitions from exploration to exploitation. This behavior indicates that the agent has progressively refined its temporal decision-making to improve its ability to identify the optimal time window for placing prediction points. 

Achieving a prediction horizon of up to 8.72 seconds for subject-independent and 7.89 seconds for subject-dependent evaluations offers major clinical benefits. This extended window provides sufficient time to accommodate device latency and patients’ neuromotor reaction, enabling them to register cues and adjust gait before a full FOG episode occurs. Such proactive intervention is critical for reducing fall risk, one of the most severe complications in Parkinson’s disease. However, occasional misplacements and imbalanced training episode may lead to false alarms and limiting reliability. Refining the agent is the key to maintain early warnings while ensuring accuracy and consistency and improving safety and effectiveness in real-world clinical applications.

%% file: Authors/6_conclusion.tex
\section{Conclusion}
\label{section:Conclusion}
This study demonstrates the agent’s ability to predict freezing of gait (FOG) episodes with horizons of up to 8.72 seconds in subject-independent evaluations and 7.89 seconds in subject-dependent evaluations. These extended prediction windows provide significant clinical benefits, particularly by enabling timely activation of cueing systems to reduce FOG-related risks. Consistent early predictions support proactive interventions that may lower fall incidents among individuals with Parkinson’s disease, although challenges such as data imbalance and occasional misclassifications remain.

Future development will focus on improving model robustness and clinical applicability through LSTM integration. The roadmap includes enriching temporal state encoding to capture complex pre-FOG gait dynamics, enabling adaptive real-time personalization, and validating the enhanced model via clinical trials. Additional strategies involve signal smoothing to reduce false alarms, a two-step confirmation mechanism, and automatic episode addition to address data imbalance. Sensitivity analysis of reward parameters will optimize prediction horizons while architectural adjustments will ensure efficiency for wearable devices—minimizing power consumption and maximizing real-time reliability.

%% file: Authors/App_A.tex
\section{Pseudocode}
\label{section:appendixA}
\begin{algorithm}[H]
\small                 
\captionsetup{skip=4pt} 
\caption{DDQN+PER Agent (Decision: \textit{wait}/\textit{place}) for FOG Prediction}
\label{alg:ddqn_per_concise}
\begin{algorithmic}[1]

\Statex \textbf{Action space:} $\mathcal{A}=\{0:\texttt{wait},\,1:\texttt{place}\}$
\Statex \textbf{State:} $s \in \mathbb{R}^6$ (6-parameter vector)

\Procedure{Train}{Total\_Episodes}
  \State $Q_{\text{network}} \gets \text{Create\_Neural\_Network}(\theta)$
  \State $Q_{\text{target}} \gets \text{Create\_Neural\_Network}(\theta' \leftarrow \theta)$
  \State $\mathcal{D} \gets \text{Initialize\_Prioritized\_Replay\_Buffer}(\text{capacity})$
  \State \textbf{Hyperparams:} $\gamma,\,\epsilon,\,\epsilon_{\min},\,\epsilon_{\text{decay}},\,\alpha,\,\beta_0,\,\beta_{\text{inc}},\,\text{batch\_size},\,C_{\text{target}}, \text{n\_steps}$

  \For{$\text{episode} \gets 1$ \textbf{to} Total\_Episodes}
    \State $s \gets \text{Environment.reset()} \quad$ \Comment{$s \in \mathbb{R}^6$}
    \State $\text{done} \gets \textbf{False}$
    \While{\textbf{not} done}
      \State \Comment{\textbf{1) Select Action} ($\epsilon$-greedy) over $\{0=\texttt{wait},1=\texttt{place}\}$}
      \If{$\text{Uniform}(0,1) < \epsilon$}  $a \gets \text{random choice in } \{0,1\}$
      \Else \quad $a \gets \arg\max_{a' \in \{0,1\}} Q_{\text{network}}(s, a'; \theta)$
      \EndIf

      \State \Comment{\textbf{2) Step Environment} and get next 6-parameter state}
      \State $(s', r, \text{done}) \gets \text{Environment.step}(a) \quad$ \Comment{$s' \in \mathbb{R}^6$}

      \State \Comment{\textbf{3) Store Transition in PER (SumTree)}}
      \State $\mathcal{D}.\text{add}(s, a, r, s', \text{done})$

      \State \Comment{\textbf{4) Learn via DDQN + PER}}
      \If{$|\mathcal{D}| \ge \text{batch\_size}$}
        \State $\beta \gets \text{Schedule}(\beta_0, \beta_{\text{inc}}, \text{n\_steps})$
        \State $(\{(s_j,a_j,r_j,s'_j,\text{done}_j)\}, \text{idxs}, \{w_j\}) \gets \mathcal{D}.\text{sample}(\text{batch\_size}; \alpha, \beta)$
        \For{$j \gets 1$ \textbf{to} \text{batch\_size}}
          \State $a^{*}_j \gets \arg\max_{a'} Q_{\text{network}}(s'_j, a'; \theta)$ \Comment{\textit{DDQN: action by online}}
          \State $y_j \gets \begin{cases}
             r_j, & \text{if } \text{done}_j\\
             r_j + \gamma \cdot Q_{\text{target}}(s'_j, a^{*}_j; \theta'), & \text{otherwise}
          \end{cases}$
          \State $\delta_j \gets y_j - Q_{\text{network}}(s_j, a_j; \theta)$ \Comment{\textit{TD error}}
        \EndFor
        \State $\text{loss} \gets \frac{1}{\text{batch\_size}} \sum_{j} w_j \cdot \delta_j^2 \quad$ \Comment{\textit{IS-weighted MSE}}
        \State $\theta \gets \text{Optimize}(\theta, \nabla_\theta \text{loss})$
        \State $\mathcal{D}.\text{update}(\text{idxs}, |\delta|)$ \Comment{\textit{refresh priorities with } $|\delta_j|$}
      \EndIf

      \State \Comment{\textbf{5) Target-Network Sync (periodic)}}
      \If{$(\text{n\_steps} \bmod C_{\text{target}}) = 0$}
        \State $\theta' \gets \theta$
      \EndIf

      \State $s \gets s'$
      \State $\text{n\_steps} \gets \text{n\_steps} + 1$
    \EndWhile

    \State \Comment{\textbf{6) Epsilon Decay per Episode}}
    \State $\epsilon \gets \max(\epsilon_{\min},\; \epsilon \cdot \epsilon_{\text{decay}})$
  \EndFor
\EndProcedure

\end{algorithmic}
\end{algorithm}

%% file: Authors/App_B.tex
\section{RL Baseline Results}
\label{section:appendixB}

\textcolor{black}{A simple reward scheme and the implementation without six statistical parameters serve as the RL baseline. The RL baseline acts as the initial model for developing the FoG prediction model in this study. The simple reward scheme consists of: +100 for accurate predictions, $-40$ for early predictions (>15 s), $-60$ for late predictions (<6 s), $-200$ for failure, but without the +0.1 reward for each `wait' action. This scheme makes the agent less motivated to make predictions (Table \ref{tab:impact} -- Simple Reward Scheme) by a large amount of undecided prediction point.}

\begin{figure}[b]
\begin{center}
\includegraphics[width=1.0\textwidth]{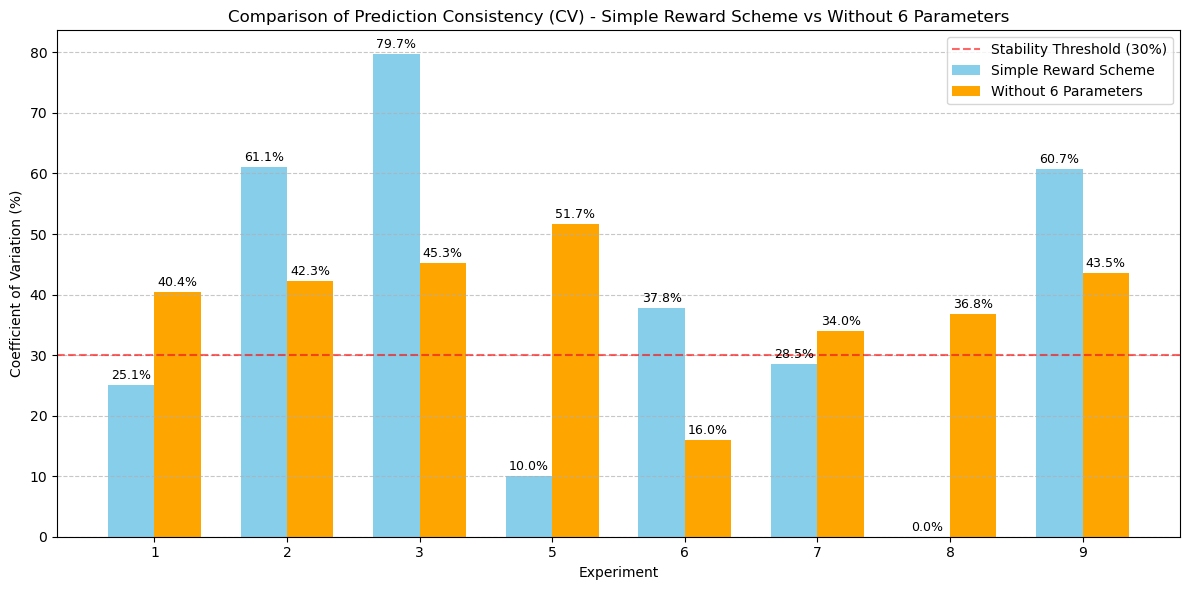}
\caption{\label{fig:stableSTD} Stability Comparison between Simple Reward Scheme vs Without 6 Statistical Parameters}
\end{center}
\end{figure}

\textcolor{black}{The implementation without the six statistical parameters serves as an ablation experiment for the agent to make predictions. The results of this ablation experiment show that the undecided prediction point decreased and the longest prediction horizon became significantly longer. Although these numerical results seem promising, a larger variance emerged in the ablation experiment, indicating less stable predictions. The Coefficient of Variation (CV) analysis shows that predictions under the Simple Reward Scheme are still better than those in the ablation experiment because they exhibit smaller variation relative to the mean (Fig. \ref{fig:stableSTD}). }

\begin{table}[h]
    \centering
    \color{black}
    \caption{Simple Reward Scheme and Ablation Experiment}
    \label{tab:impact}
    \resizebox{\textwidth}{!}{%
    \begin{tabular}{l *{8}{c}} 
        \toprule
        \textbf{Variable} & \multicolumn{8}{c}{\textbf{Subject ID}} \\
        \cmidrule(lr){2-9}
        & \textbf{1} & \textbf{2} & \textbf{3} & \textbf{5} & \textbf{6} & \textbf{7} & \textbf{8} & \textbf{9} \\
        \midrule
        \multicolumn{9}{l}{\textbf{Simple Reward Scheme}} \\
        \addlinespace[2pt] 
        Mean - reward & 74.28 & -49.34 & 0.17 & 145.05 & 82.23 & 96.45 & 151.50 & 6.41 \\
        Mean -  prediction point (s) & 5.85 & 5.76 & 4.18 & 5.71 & 4.71 & 6.39 & 4.62 & 5.91 \\
        Std Dev - \textcolor{black}{prediction point (s)} & 1.47 & 3.52 & 3.33 & 0.57 & 1.78 & 1.82 & 0.00 & 3.59 \\
        The longest prediction horizon (s) & 8.49 & 10.16 & 11.63 & 6.24 & 6.19 & 11.72 & 4.62 & 14.35 \\
        The shortest prediction horizon (s) & 0.40 & 2.34 & 0.39 & 1.71 & 2.21 & 1.56 & 4.62 & 1.43 \\
        Misplaced ratio & 1:19 & 1:23 & 2:40 & 8:58 & 1:10 & 1:21 & 0:14 & 0:25 \\
        Undecided prediction point & 10 & 17 & 30 & 34 & 6 & 12 & 13 & 14 \\
        \addlinespace[5pt] 
        \multicolumn{9}{l}{\textbf{Without 6 Parameters}} \\
        \addlinespace[3pt]
        Mean - reward & 96.28 & 50.08 & 63.68 & 16.97 & -39.58 & 22.53 & 55.96 & -25.99 \\
        Mean - \textcolor{black}{prediction point (s)} & 7.01 & 7.55 & 5.41 & 6.67 & 12.70 & 7.20 & 8.17 & 7.61 \\
        Std Dev - \textcolor{black}{prediction point (s)} & 2.83 & 3.19 & 2.45 & 3.45 & 2.03 & 2.45 & 3.01 & 3.31 \\
        The longest prediction horizon (s) & 14.95 & 14.84 & 14.96 & 14.72 & 14.58 & 14.84 & 11.89 & 12.91 \\
        The shortest prediction horizon (s) & 1.62 & 3.91 & 0.42 & 0.89 & 9.72 & 2.34 & 4.62 & 1.91 \\
        Misplaced ratio & 1:19 & 2:22 & 6:41 & 8:57 & 1:10 & 2:21 & 4:14 & 0:25 \\
        Undecided prediction point & 7 & 11 & 15 & 28 & 5 & 4 & 2 & 13 \\
        \bottomrule
    \end{tabular}
    } 
\end{table}

